\pdfoutput=1

\documentclass[11pt]{article}

\usepackage[final]{EMNLP2023}

\usepackage{times}
\usepackage{latexsym}
\usepackage{stfloats}  

\usepackage[T1]{fontenc}

\usepackage[utf8]{inputenc}

\usepackage{microtype}

\usepackage{inconsolata}

\usepackage{booktabs}

\usepackage[acronym]{glossaries}  
\usepackage{graphicx}
\graphicspath{ {./images/} }

\usepackage{subcaption} 

\usepackage{float}

\usepackage{algorithm}      
\usepackage{algpseudocode}
\usepackage{multirow}
\usepackage{amsthm}
\usepackage{amsfonts}
\usepackage{cleveref}
\crefname{lstlisting}{listing}{listings}
\Crefname{lstlisting}{Listing}{Listings}
\usepackage{accsupp} 
\usepackage{listings}
\usepackage{hyperref}
\definecolor{beige}{rgb}{0.96, 0.96, 0.86} 
\newcommand{\TESTPROMPT}{\makebox[0pt][l]{\color{beige}\rule[-3pt]{\linewidth}{11pt}}}

\theoremstyle{definition}
\newtheorem{definition}{Definition}[section]

\newacronym{llm}{LLM}{large language model}
\newacronym{lm}{LM}{language model}
\newacronym{nlp}{NLP}{natural language processing}
\newacronym{mlp}{MLP}{multilayer perceptron}

\title{Self-Consistency of Large Language Models under Ambiguity}

\author{%
    Henning Bartsch* \\
  Independent Researcher \\
  \texttt{bartsch.henning@gmail.com} \\
   \\
  \And
   Ole Jorgensen* \\
  Imperial College London \\
  \texttt{okj22@ic.ac.uk} \\
   \\
  \And
  Domenic Rosati* \\
  Dalhousie University\\
  \texttt{domenic.rosati@dal.ca} \\
 \AND
  Jason Hoelscher-Obermaier \\
  PIBBSS fellow \\
\texttt{jason.hoelscherobermaier@gmail.com} \\
   \\
 \And
  Jacob Pfau \\
  New York University \\
  \texttt{jp6263@nyu.edu}
}

\begin{document}
\maketitle
\begin{abstract}

Large language models (LLMs) that do not give consistent answers across contexts are problematic when used for tasks with expectations of consistency--e.g. question-answering, explanations, etc. Our work presents an evaluation benchmark for self-consistency in cases of under-specification where two or more answers can be correct. We conduct a series of behavioral experiments on the OpenAI model suite using an ambiguous integer sequence completion task. We find that average consistency ranges from 67\% to 82\%, far higher than would be predicted if a model's consistency was random, and increases as model capability improves. Furthermore, we show that models tend to maintain self-consistency across a series of robustness checks, including prompting speaker changes and sequence length changes. These results suggest that self-consistency arises as an emergent capability without specifically training for it. Despite this, we find that models are uncalibrated when judging their own consistency, with models displaying both over- and under-confidence. We also propose a nonparametric test for determining from token output distribution whether a model assigns non-trivial probability to alternative answers. Using this test, we find that despite increases in self-consistency, models usually place significant weight on alternative, inconsistent answers. This distribution of probability mass provides evidence that even highly self-consistent models internally compute multiple possible responses.
\end{abstract}

\section{Introduction}

Language model pre-training approximates a distribution generated by many speakers. As a result, LLMs learn to express inconsistent beliefs drawn from distinct groups of people \cite{santurkar2023opinions}. Recent work has investigated the consistency of LLMs variously as: a logical validity check on model claims \cite{fluri2023evaluating}, an explanatory validity check on the simulatability of models’ explanations \cite{chen2023models}, and a tool to identify LLMs representations of truth \cite{burns2022discovering}. All of these works rest to some degree on the contention that fine-tuned LLMs can be understood as holding beliefs, an assumption which has recently come under scrutiny \cite{levinstein2023lie}. 

Consistency is particularly of interest in cases of ambiguity. Recent work has evaluated LLMs' ability to identify linguistic and classification-task ambiguity \cite{liu2023were, tamkin2022task}. Our work brings together these threads of research, examining how model explanations can be examined via self-consistency checks. 

We offer a case study on ambiguity in an arithmetical setting. We ask language models from OpenAI for a \textit{continuation} of an integer sequence having multiple possible continuations. We then separately ask the models for the formula that generated the initial sequence, which we refer to as the \textit{explanation}. Finally, we evaluate whether model-generated \textit{continuations} are consistent with model-generated \textit{explanations} (§\ref{q0_capability_consistency}). We present the model with the full set of sequence generating functions so that ambiguity is, in principle, recognizable by the model.




We find the following across evaluations using \texttt{davinci} (GPT-3), \texttt{text-davinci-003}, \texttt{gpt-3.5-turbo}, and \texttt{gpt-4}:
\begin{enumerate}
    \item Models (with greedy decoding) improve in cross-context consistency rapidly with increasing scale and capabilities (§\ref{q1_self_consistency_changes}). This holds across prompting strategies and data perturbations (§\ref{q1_speaker_changes}).
    \item Models are not well-calibrated and incapable of self-assessing the consistency of their own answers (\Cref{fig:consistency}). 
    \item Even a model (\texttt{text-davinci-003}) that chooses relatively consistently among several correct answers across contexts still assigns non-trivial probability to other correct answers (§\ref{q2_alternative_considerations}).
    \item Models can generally verbalize alternative answers in cases of ambiguity, but there is no clear effect of capability increase on this verbalization task (§\ref{q2_verbalize}).
\end{enumerate}



\section{Dataset: Ambiguous Integer Sequences}
\label{dataset_int_seq}

In order to evaluate self-consistency, we created and open-sourced a dataset of ambiguous integer sequences.\footnote{\href{https://github.com/JacobPfau/introspective-self-consistency}{https://github.com/JacobPfau/introspective-self-consistency}}. Integer sequences were chosen because we can readily identify sequences that have multiple valid completions. This allows us to introduce tasks with ambiguity for measuring properties like model self-consistency. Previous work on self-consistency considered open-ended question answering or knowledge probing \cite{raj2023measuring, elazar-etal-2021-measuring} which makes measuring consistency difficult (rendering unclear the space of possible answers, and what constitutes distinct answers), whereas in our setting the space of possible answers is rigorously defined via an enumeration of generation functions.

Our dataset was created as follows: We generate integer sequences, e.g., $7, 11, 15$, drawn from a fixed set of generating functions, e.g., $\texttt{lambda x: (4 * x) + 3}$. \Cref{tab:integer_sequence_example} illustrates some examples drawn from our dataset. The underlying function is referred to as the \textit{rule} or \textit{explanation} of the sequence, and the next integer as the \textit{completion}. Our experimental settings are mostly based on two fundamental tasks: (1) sequence completion and (2) sequence explanation. For completions, we query models for the next item in a given integer sequence. For explanations, models are prompted for the underlying function that generated the given sequence. In our experiments, models should return explanations in the form of Python lambda functions whose form is demonstrated through few-shot examples (see \Cref{app:prompt_examples}). Models are informed of the function space ahead of time by being presented with the possible generating functions in the instruction prompt.

\begin{table}[]
\centering
\small
\begin{tabular}{lll}
\toprule
\multicolumn{1}{c}{Sequence}     & \multicolumn{1}{c}{Completion} & \multicolumn{1}{c}{Rule}                   \\ 
\midrule
4, 6, 8 & 10      & \texttt{lambda x: x + 2} \\
7, 11, 15    & 15         & \texttt{lambda x: (3 * x) | 3}  \\
7, 11, 15    & 19         & \texttt{lambda x: (4 * x) + 3}  \\ 
\bottomrule
\end{tabular}
\caption{\label{tab:integer_sequence_example} Example of integer sequences that are either unambiguous or ambiguous given a specific set of generating rules (enumerated in \Cref{tab:function_templates}).}
\end{table}

\textit{Ambiguous} sequences are sequences for which there are multiple generating rules which differ in their continuation of the sequence. Unambiguous sequences are sequences which have only one valid completion within our function space. \Cref{app:mining_ambigious_sequences} describes our algorithm for mining for ambiguous sequences as well as the parameters of the function space we searched over. The function space consists of eight function templates, each with two constant arguments. We generate functions from those templates by setting the constant terms in the range $[0,4]$, resulting in 197  possible functions on which \Cref{alg:mining_ambigious_sequences} is used. Our dataset consists of 140 unambiguous sequences and 57 ambiguous sequences. 

\section{Methodology: Evaluating Consistency}
\label{q0_capability_consistency}

We measure consistency by comparing responses from the completion task to responses from the explanation task, which we call \textit{cross-context} because the model sees each task in a separate context window. Each prompt uses eight demonstrations showing the model how to complete the sequence or explain the sequence using a Python function. The demonstrations are drawn randomly\footnote{To control for the effect of these random sequences on biasing consistency, we report results aggregated from multiple runs} from the same function space as the ambiguous and unambiguous functions. Examples of these prompts are presented in \Cref{app:prompt_examples}.

The models chosen for evaluation were \texttt{text-davinci-003}, \texttt{gpt-3.5-turbo}, and \texttt{gpt-4}.\footnote{\href{https://platform.openai.com/docs/models.}{https://platform.openai.com/docs/models}. For \texttt{gpt-4}, we use the \texttt{gpt-4-0314} version. For \texttt{gpt-3.5-turbo}, we use the model that was available from March to June 2023.} While we are not entirely sure how these models are trained, these models were chosen because they are commonly used by both researchers and the public, and they represent a sequence of capability increases through data quality improvement, annotations, and innovation in training and inference techniques (see \citet{openai2023gpt4}).

In the below experiments, \textit{greedy sampling} (temperature set to $0$) is used throughout. This choice lets us conduct a best-case analysis of self-consistency: studying whether a model is \textit{capable of} self-consistency when the sampling strategy is advantageous. In §\ref{q2_alternative_considerations}, we move on from greedy decoding and examine what the full output distribution implies about the possible continuation space of models. \footnote{Given the nature of black-box API-based evaluation, it is possible greedy decoding doesn't ensure determinism (e.g. because of sparse mixture of experts routing considerations).}

\subsection{Explanation and completion accuracy}
\label{q0_capability}

Before considering cross-context consistency, we first benchmark these models' accuracy on sequence completion or sequence explanation in unambiguous cases. For the completion case, we present the models with a sequence of four integers and evaluate its accuracy on generating the next item in the sequence. For the explanation case, we present the models with a sequence and evaluate the model's accuracy on generating an exact match of the Python function used to produce the sequence.

\begin{table}[h]
\centering
\small
\begin{tabular}{llll}
\toprule
                      & \multicolumn{2}{c}{Accuracy (\%)} & \% \\
Model                 & Explanation & Completion & Valid \\
\midrule
\texttt{davinci}          & 6.00  & 20.20 & 95.5 \\
\texttt{text-davinci-003} & 31.18  & 65.95 & 99.3 \\
\texttt{gpt-3.5-turbo}    & 50.25  & 77.56 & 97.6 \\
\texttt{gpt-4}       & 59.05  & 78.64 & 94.8 \\
\bottomrule
\end{tabular}
\caption{\label{tab:capability} Mean explanation and completion accuracy scores in unambiguous cases, as well as fraction of valid, parseable answers, for each model across three runs. Accuracy increases with general model capability and is higher for completion than for explanation.}
\end{table}

\Cref{tab:capability} presents our capability results. We report the average explanation and completion accuracy scores across three runs. We also report the fraction of valid answers (out of a total of 140 test cases, our unambiguous functions) where the model provided a valid parseable answer, such as a valid integer or Python function. The results are largely intuitive: as general model capacity increases, performance on the explanation and completion tasks increases. Note that the explanation task is generally harder than the completion task. On both tasks, \texttt{davinci} does poorly despite having a high number of valid answers, so \texttt{davinci} was not used in subsequent experiments.

\subsection{Explanation and completion consistency}
\label{q0_consistency}

\begin{figure*}[]
\centering
\includegraphics[scale=0.36]{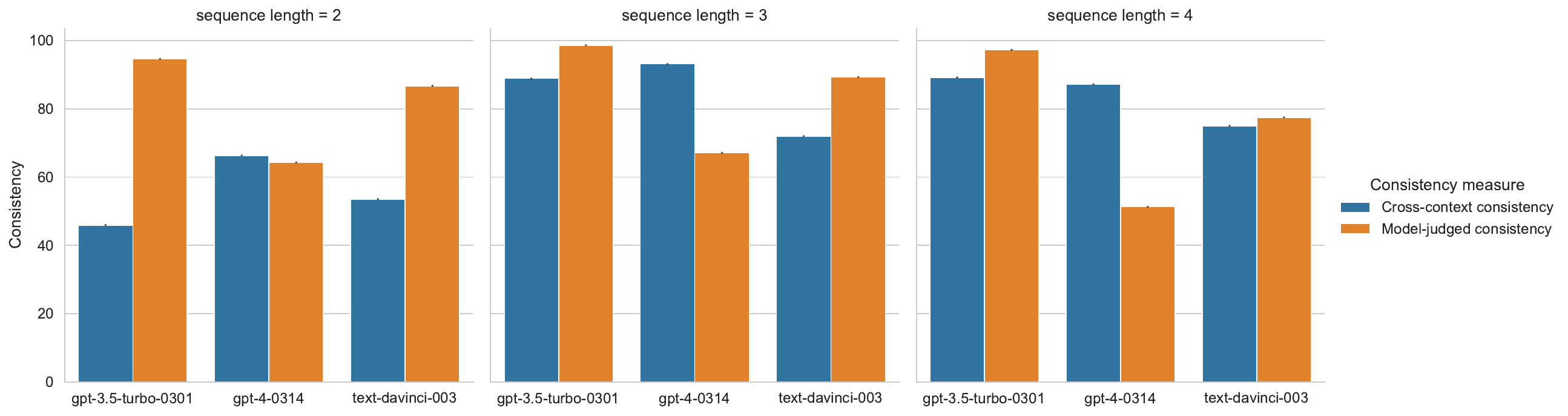}
\caption{Cross-context consistency (orange). Model-judged consistency (blue); this drops drastically for \texttt{gpt-4}, which underestimates the consistency across answers itself produced.}
\label{fig:consistency}
\end{figure*}

Our second set of experiments evaluates the consistency of a given explanation for a sequence and a completion for the same sequence when a model is prompted separately for explanation and completion. We use a similar setup as the previous experiment, including the explanation and completion prompts used earlier. We measure the following (see \Cref{app:prompt_examples} for corresponding prompts):
\begin{itemize}
  \item \textbf{Cross-context consistency:} whether the explanation provided by the model generates the given sequence, including the completion provided separately by the model. 
  \item \textbf{Model-judged consistency:} whether the model, itself, judges the explanation (rule) it provided and the completion it provided to be consistent, i.e., the rule generates the sequence with claimed completion (see \Cref{lst:prompt_self_consistency} for the prompt used in these judgements).
\end{itemize}

\Cref{fig:consistency} illustrates the performance of each model on the above scores when we vary the number of integers in the initial sequence from a length of two to a length of four. Sequences with two initial integers have 196 ambiguous sequences, three initial integers has 76 total ambiguous sequences, and four initial integers have 140 ambiguous sequences. This variance allow us to understand the behavior of models as the space of ambiguity varies. The two main results are (1) model improve in consistency as they improve in arithmetical capability from \texttt{text-davinci-003} to \texttt{gpt-4}, (2) models tend to consider their answers consistent when they are not, except for \texttt{gpt-4} which underestimates its own consistency. Result (2) is noteworthy because calibration, or the ability of a model to express accurate estimates of its own behavior, is an important safety property of LLMs \cite{fluri2023evaluating, lin2022teaching}. In domains where human evaluation cannot be done, \citet{fluri2023evaluating} identify model self-evaluations of consistency as a primary method useful for invalidating untrustworthy responses. A well-calibrated model should have \textit{cross-context consistency} and \textit{model-judged consistency} scores as close as possible.

\subsection{Consistency and Capability}
\label{q0_correlation}
\begin{figure*}[h]
\centering
\includegraphics[scale=0.6]{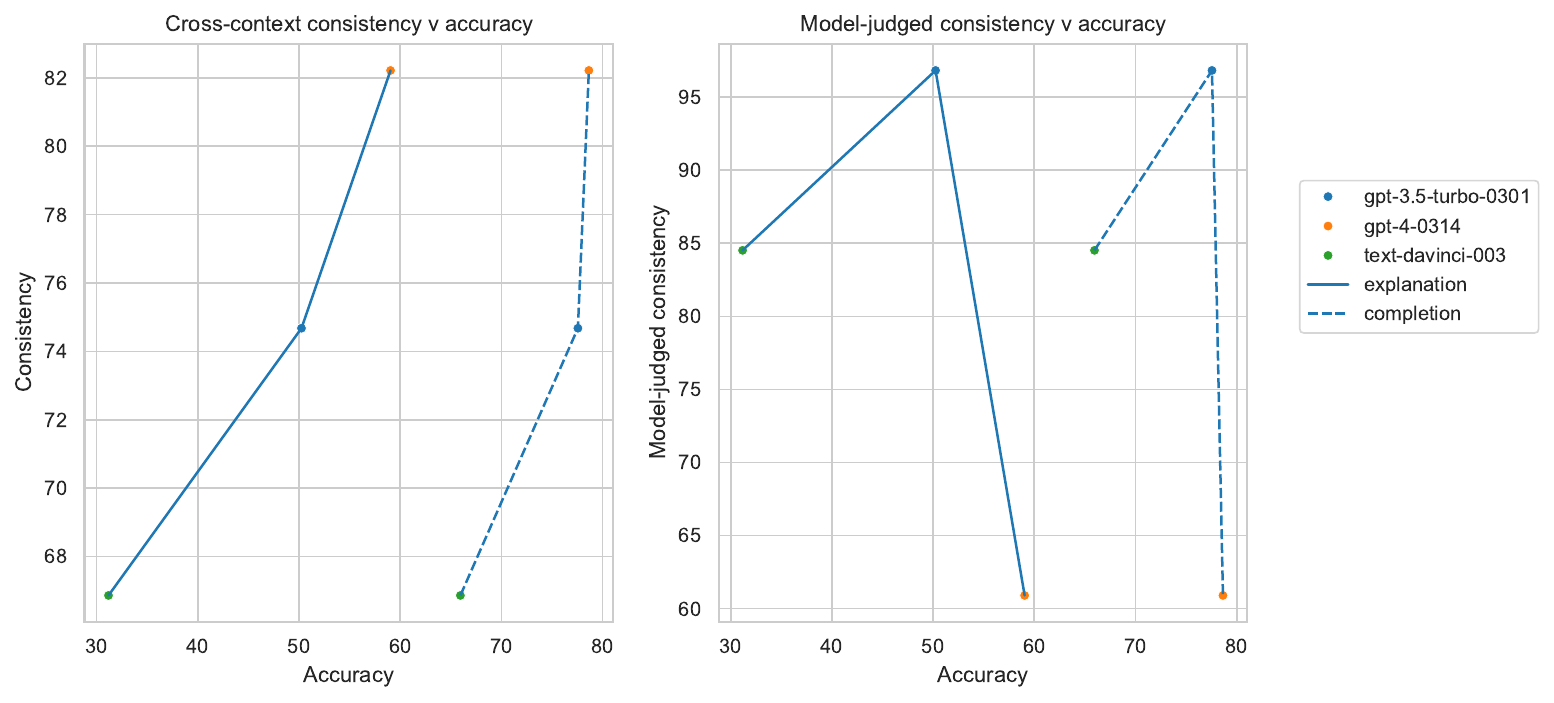}
\caption{
 Explanation and sequence completion accuracies plotted against cross-context consistency and model-judged consistency (mean over sequence lengths). Further illustration of \texttt{gpt-4}'s inability to correctly assess its own consistency despite being much more consistent.
}
\label{fig:correlation}
\end{figure*}

\Cref{fig:correlation} presents the results from §\ref{q0_capability} and §\ref{q0_consistency} plotted together. This analysis investigates the degree to which model capability relate separately with \textit{cross-context consistency}, and \textit{model-judged consistency}. We see as capability increases so does \textit{cross-context consistency} but, the most capable model \texttt{gpt-4} is worse evaluating its own consistency.

Additionally, we compute expected consistency if correct completion-explanation pairs were chosen uniformly randomly at different capability thresholds. \Cref{tab:expected_consistency} illustrates cross-context consistency performance by our models and expected random consistency based on the average performance of each model on explanation and sequence completion accuracy. This tells us how consistent we should expect models to perform at different capability levels if they chose their completion responses independently from their explanations. Note that a model could score perfectly on the capability evaluations and consistency evaluations while having no self-consistency whatsoever. What we find is that models approach perfect consistency rapidly with capability increases. 


\begin{table}[h]
\small
\centering
\begin{tabular}{lrr}
\toprule
        & \multicolumn{2}{c}{Average consistency (\%)} \\
Model         & Actual & Random \\ 
\midrule
\texttt{text-davinci-003} & 66.86  & 8.50 \\
\texttt{gpt-3.5-turbo}       & 74.68  & 10.02 \\
\texttt{gpt-4}    & 82.22  & 15.22 \\
\bottomrule
\end{tabular}
\caption{\label{tab:expected_consistency} Average cross-context (Actual) consistency across settings in \Cref{fig:consistency} and consistency we'd expect to see (Random) if valid answers were selected uniform randomly given the average accuracy performance for each model.}
\end{table}

\section{Robustness Checks for Consistency}
\label{q1_self_consistency_changes}


\begin{figure*}[h]
    \begin{subfigure}{0.33\textwidth}
        \centering
        \includegraphics[width=\linewidth]{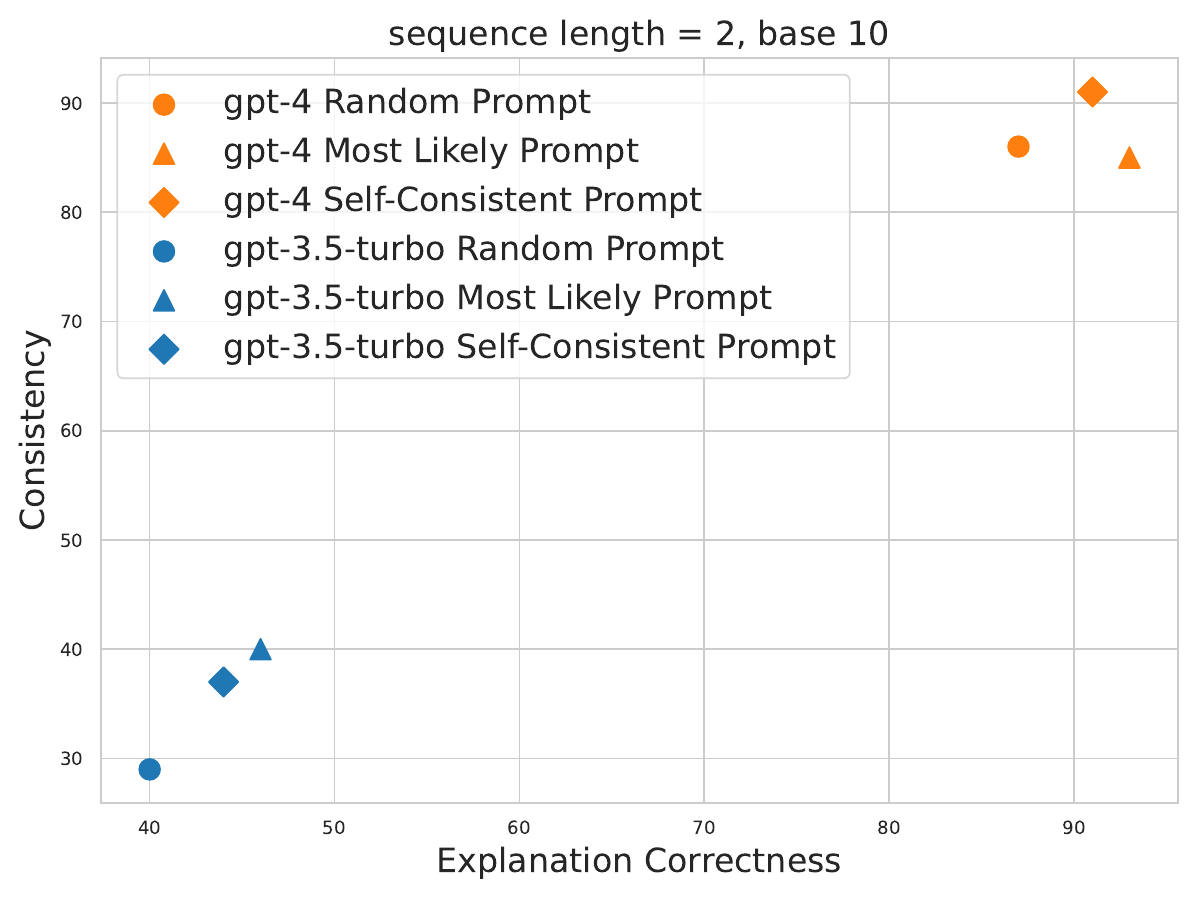}
        \label{fig:speaker-graph}
    \end{subfigure}
    \hfill
        \begin{subfigure}{0.66\textwidth}
        \centering
        \includegraphics[width=\linewidth]{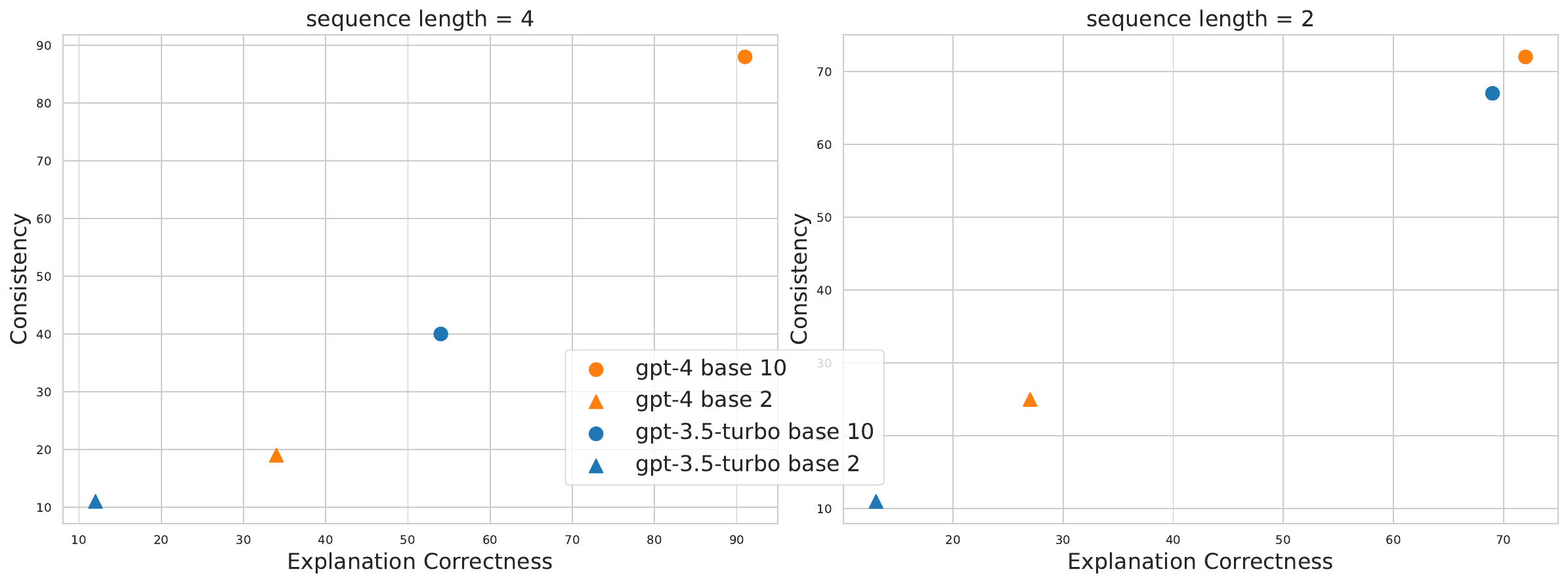}
        \label{fig:base-change}
    \end{subfigure}
    \caption{Cross-context consistency plotted against explanation correctness, varying either the role prompt (left-hand side) or the base-representation of the integer sequences being evaluated on (middle and right-hand side).}
    \label{fig:robust-graphs}
\end{figure*}

We conducted further experiments to better understand how robust these results were to changes in experimental protocol by using a range of different prompts. We consider: (1) \textit{speaker changes} in which we prompt the model as if the sequences were generated by different speakers; (2) \textit{change of base} in which the sequence integers are presented in base 2 instead of base 10; and (2) \textit{sequence length changes}. Full results are given in §\ref{sec:robustness-results}.



\subsection{Consistency Across Speaker Changes}
\label{q1_speaker_changes}

The first robustness experiment was designed to investigate the robustness of self-consistency of models when asked to simulate different speakers. This was intended to investigate whether models could be prompted to simulate more or less self-consistent speakers, which would determine whether models should be expected to be self-consistent by default or whether the previous results were artifacts of arbitrary features of the prompt.

To do this, we again conducted the same experiments as §\ref{q0_capability_consistency}, now varying the initial instruction given to the model. These instructions were split into two separate components which we varied independently: what task we wanted the model to complete, and which speaker we wanted the model to simulate completing that task. We used three different task prompts, which one might expect to correspond to three different levels of consistency: the self-consistent prompts asked explicitly for a pair of responses which matched each other; the most likely prompts asked for the most likely continuation / explanation (most likely); and the random prompt asked the model to choose responses randomly when there was ambiguity about the correct answer. The prompts in full can be found in \Cref{app:prompt_examples}. For example, the random explanation prompt was "Assume the sequence is generated by some deterministic function. If multiple functions could generate the sequence, choose the corresponding continuation randomly".


The first plot in \Cref{fig:robust-graphs} shows representative results when varying the task prompt on correctness and consistency. If the models were capable of computing multiple continuations, and merely appeared self-consistent by dropping other possibilities, then we might expect there to be variable self-consistency, e.g., higher on the self-consistency prompt, and lowest on the random prompt. Empirically, we found that prompting the models with these different tasks had little influence on the proportion of answers that were self-consistent. This was found both for sequences of length $4$ and $2$. Even in the case where we were able to elicit a high proportion of correct answers being inconsistent using the most likely prompt, we do not see large changes in the number of inconsistent responses when varying the task prompt. This serves as strong evidence that the relationship between capability and consistency is unaffected by task prompt.

\subsection{Consistency Across Base Changes}
\label{q1_base_changes}

In this robustness experiment, we investigate what impact the base representation of functions and sequences had on capabilities and consistency of the models. This was intended to investigate the relationship between model capability and self-consistency while holding model type and training constant. We hypothesised that bases besides base $10$ would be more difficult for the model. We again prompted the model to produce a continuation of a sequence and an explanation for the sequence, although the sequences were now in base $2$, and the functions were expected to output base $2$ representations of integers.

 The second plot in \Cref{fig:robust-graphs} presents a correlation analysis for this experiment, considering both base $10$ and base $2$ responses. It demonstrates a very strong correlation between the model generating correct explanations and being self-consistent, suggesting that this trend is robust across bases and, thus, task difficulty.

\section{Distributional Analysis of Model Consistency}
\label{q2_alternative_considerations}
\subsection{Models Do Not Converge to Calculating a Unique Solution}
In the analysis so far, greedy sampling was used throughout. We now pose a follow-up question: Given models increasingly converge to self-consistency, preferring a unique answer, to what extent do these models calculate representations of other alternative answers? And, when models place high probability on alternative answers, can they verbalize these alternative solutions serially?



Specifically for models that were fine-tuned with RLHF \cite{christiano2017RLHF, ouyang2022training}, the output probabilities may not be well-calibrated to the relative frequency of tokens if the objective of RLHF encourages models to allocate probability mass narrowly \cite{kadavath2022lmknow}. Hence, models' token probability distribution may not be reflective of their credences. While the models may be uncalibrated, we make a weaker assumption below that model output probabilities are \textit{nonparametrically calibrated}: higher probability mass implies higher credence. 

Applying this assumption to our setting, given initial ambiguous sequence, $S_n$, generating rules $\{F\}$, we can determine whether a model has calculated an alternative correct sequence completion, $c'$, other than the modal greedy-decoded solution by verifying that:

\begin{equation} \label{eq:ineq}
P(c'|S_n) > P(z|S_n) \text{ for all } z \in \mathbb{N} \setminus C
\end{equation}

where $C$ is the set of correct continuations of $S_n$ and $\mathbb{N}$ is the set of all continuations.

\begin{figure}[h]
\centering
\includegraphics[width=\columnwidth]{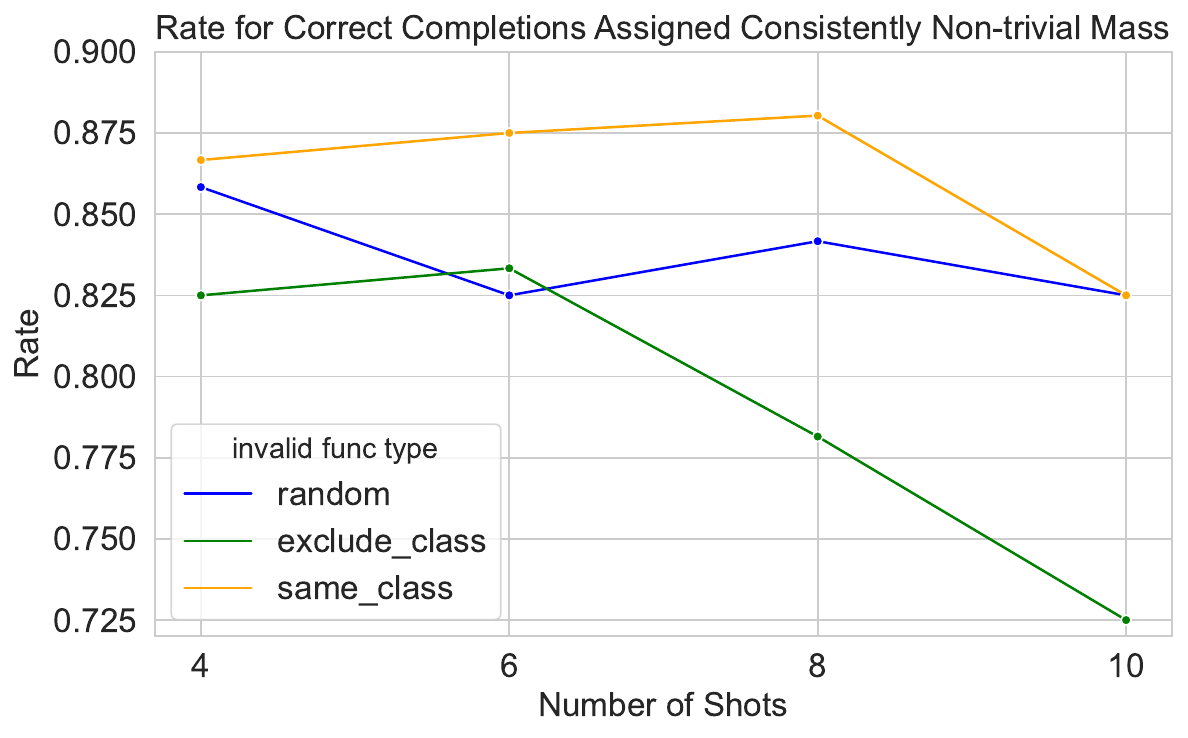}
\caption{
    Rate at which correct completion alternatives are assigned non-trivial probability mass by function class sampled for few shot exemplars. Across sampling methods, that rate is relatively high indicating a consistent consideration of correct alternatives across contexts.
}
\label{fig:ineq_test}
\end{figure}

For input data, we use the full set of 40 functions that generated ambiguous sequences (see \Cref{alg:mining_ambigious_sequences}). We prompt the model using the same prompts for integer sequence completion as in §\ref{q0_consistency} and determine whether alternative correct answers rank higher than all incorrect answers. In the explanation case, we change the prompt to be a multiple-choice task so that only a single token is needed to evaluate the above inequality. Despite this simplification, the rate at which high probability mass is spread on alternatives is much lower, with the best rate of 0.3. This indicates that correct alternatives are not generally considered. This may be because the computation of correct alternative explanations is much more computationally intensive and more difficult than the computation of correct alternative sequence completions.

We use \texttt{text-davinci-003} for our experiments since it is the only model that has token log probabilities accessible from the public API.\footnote{\href{https://platform.openai.com/docs/api-reference/completions/create}{https://platform.openai.com/docs/api-reference/completions/create}} Since the API returns up to $n_{logprobs} = 5$ probabilities for top output tokens, we assess if any incorrect answer was listed and whether the correct all rank higher. When a possible correct answer is not in the top output tokens but an incorrect one is, we consider the test failed.
Finally, we control the sampling methods for few-shot example: \texttt{exclude\_class} indicates that we exclude the sequence generating functions that are from the same class (See classes used here \Cref{tab:function_templates}), \texttt{same\_class} draws functions from the same function class and \texttt{random} draws those randomly across function classes. These controls are designed to give us insight on whether the class of functions used makes considering correct answers over incorrect ones more challenging. The evaluations are averaged over three runs.

\Cref{fig:ineq_test} illustrates that in the sequence completion case, \texttt{text-davinci-003} almost always assigns high probability to correct alternative answers. We only see small differences with function class used for few-shot examples where the cases of \texttt{same\_class} and \texttt{random} functions appear to help with computing correct alternative explanations as the number of few-shot demonstrations is increased. Sampling examples with \texttt{exclude\_class} seems to make it more challenging likely because functions that explain the model completion have not been seen before.\footnote{Since we do not have access to the underlying pre-training corpora distribution of the model, we cannot definitively rule out higher probability mass being assigned to sequences due to their frequency in the pre-training corpora.}

\begin{figure}[h]
\centering
\includegraphics[width=\columnwidth]{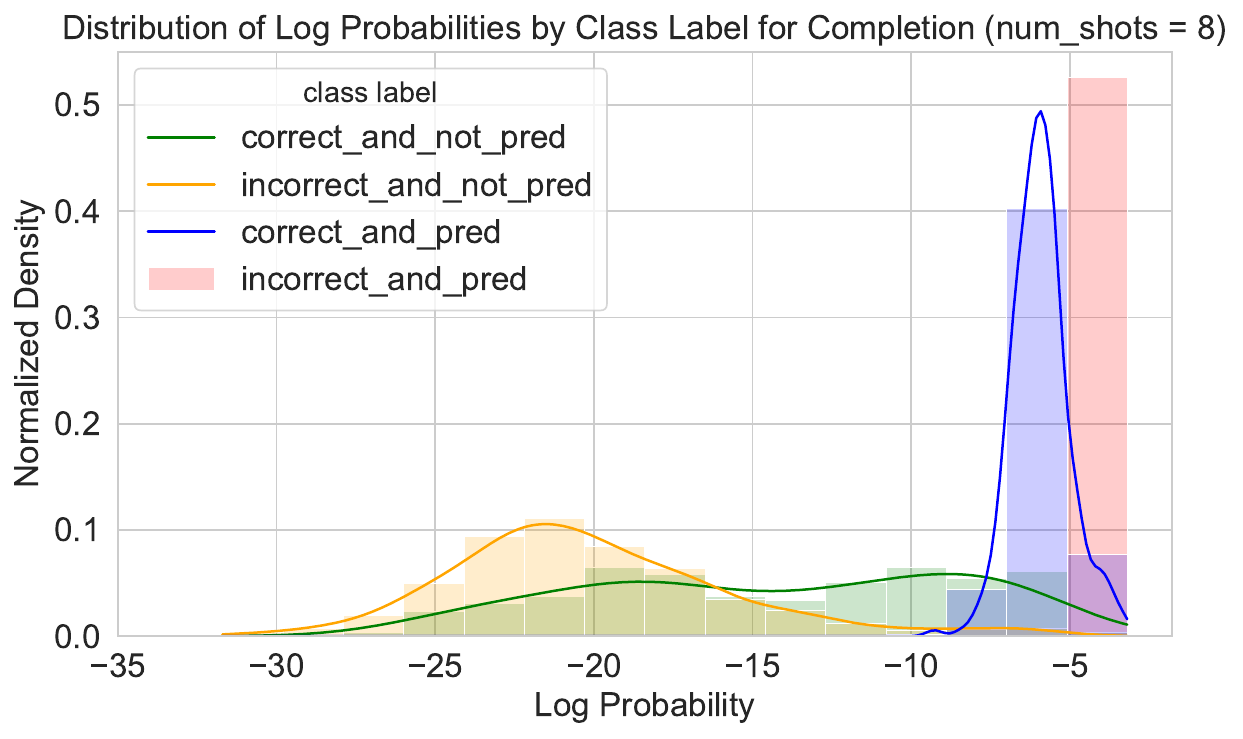}
\caption{
     Distribution over output probabilities for correct and incorrect completions for the sampling function type \texttt{random\_class}. Each histogram is normalized by the data points of the corresponding class label. With KL-divergences of $KL(\text{correct\_and\_pred} ||  \text{correct\_not\_pred})=1.71$ and $KL(\text{correct\_and\_pred} ||  \text{incorrect\_not\_pred})=3.45$ bits, the distributions of correct answers have higher overlap.\footnotemark
}
\label{fig:prob_mass}
\end{figure}
\footnotetext{To calculate the KL-divergence, we first obtained the density histograms for the same points $n_{bins} = 40$ between the minimum and maximum value of log probabilities. Additionally, we applied Gaussian smoothing with $\sigma = 1$ to include information where the quotient would otherwise have been undefined.}

In \Cref{fig:prob_mass}, the distribution over log probability mass is shown for the sequence completion task across four combinations over two variables: correctness and (greedy) prediction, i.e., whether the response in question was predicted as the top-1 response. The distribution for predicted answers look similar: correct and predicted answers (blue) narrowly concentrate relatively large log probabilities and a single peak for incorrect predictions (red). For non-predicted answers, the distributions are generally flatter and their mean shifted towards comparatively smaller values. 




For correct and non-predicted answers (green), the distributions' median at around -13.8 is much larger than at -20.7 for incorrect answers. This difference indicates that the model allocates non-trivial probability mass to those correct options. Correct alternatives are calculated and represented by the model internally. When normalizing the distribution across all data points the probability mass place on correct answers is relatively large and narrow, even for non-predicted answers (see \Cref{fig:app-prob-mass-normalized}).


\subsection{Verbalizing Alternatives}
\label{q2_verbalize}
While inspecting the probability distribution over answers gives insights into the potential consideration of alternatives, we are further interested in the extent to which models would verbalize those alternatives if prompted. This is important because outside of our simple sequence modeling cases, natural language questions will generally have distinct answers which require multiple tokens to express, making it impractical to directly read off answer probability from logits.

In this experiment, we prompt the model to provide all possible answers for an ambiguous sequence task and compare those with the correct options (prompt in \Cref{lst:prompt_verbalize_alternative_continuations}). We provide in-context examples and consider only up to $5$ alternatives. Precision and recall scores are calculated, comparing verbalized answers with the valid continuations. For input data, we consider the default ambiguous sequences (see \Cref{alg:mining_ambigious_sequences}).

\begin{figure}
    \centering
    \includegraphics[width=\columnwidth]{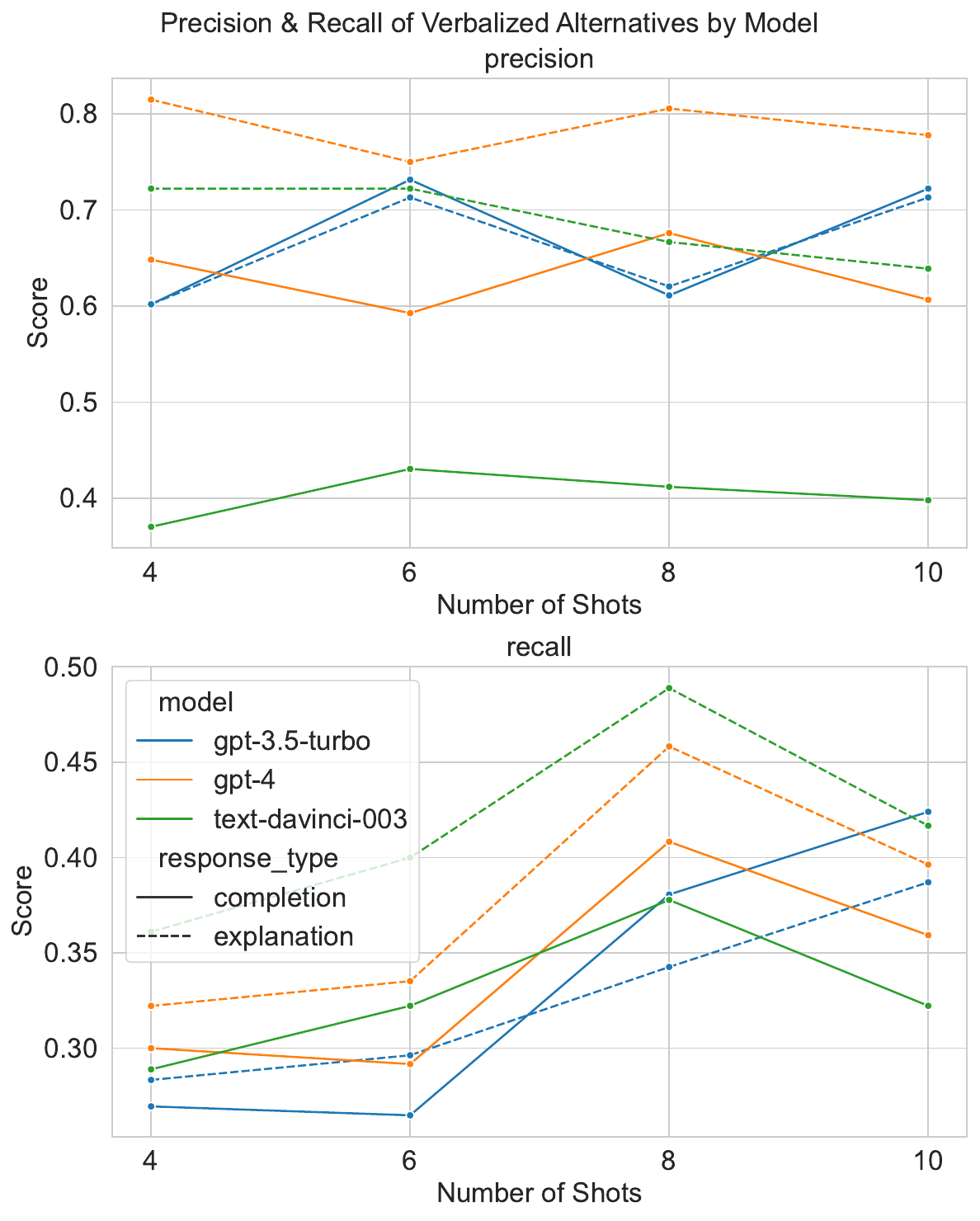}
    \caption{Precision and recall scores of alternative answers verbalized by different models compared to correct answers, up to $5$ alternatives and only distinct values were counted.}
    \label{fig:verbalize-precision}
\end{figure}

The high precision scores in \Cref{fig:verbalize-precision} show that models do not tend to produce random, incorrect answers. Recall scores are much lower, for completion reaching a maximum of 0.41 and for explanation 0.49. Compared to precision this aligns with our expectations that verbalizing \textit{all} alternatives is very difficult. However, the rapid increase in recall with additional in-context examples implies that the models adapt to include more correct alternatives.\footnote{For $n_{shots} > 10$, our prompt exceeds the token limit. Despite increasing recall scores, we were not able to investigate the impact of few-shot examples further.} In contrast to our previous results, the performance for the explanation tasks is similar to completion. \texttt{text-davinci-003} achieves the highest recall for explanation despite being the generally less capable model, but likely preserving a wider options space and multiple possible continuations due to less RLHF fine-tuning. The low precision score indicates that it thereby also produces false negatives. The relatively high recall of \texttt{gpt-4} for explanation and completion tasks show its verbalization capabilities. However, in the easier  completion task, high recall scores would be expected if the model considered more alternatives.


\section{Related Work}


Our work is motivated by previous research on truthfulness. Approaches like \citet{lin2022truthfulqa} directly tackle this problem by developing benchmarks for truthfulness of LLMs across a range of questions such as health, law, and politics. Detecting inconsistencies is helpful, but not sufficient, for evaluating the truthfulness of language models.


Evaluating model behavior under ambiguity would shed some light on this question, as explored in \citet{liu2023were}. Here, however, the emphasis is on interpreting ambiguous natural language sentences correctly, as opposed to making the same judgment in a range of different contexts. This means that failings might not be indications of inconsistency but rather a poor understanding of natural language.

Similarly, the approach towards consistency evaluations taken by \citet{fluri2023evaluating} focuses on whether different answers are logically consistent. When a set of conditions over different inputs holds, then conditions over corresponding outputs should logically follow. For instance, forecasting world records in 100m sprint should monotonically decrease over time. In contrast to our own work, the investigations focus on scenarios without known ground truth. Our focus on being consistent across contexts tests for poor world models and extends consistency checks to arithmetic reasoning tasks.

\citet{tamkin2022task} presents a novel benchmark for studying how well models are able to detect salient features of sentences where this salient feature is undetermined. This relates to our ambiguous sequences setting, although the focus on interpreting natural language means the evaluations will not separate poor language understanding from inherent inconsistency.

Self-consistency also relates to chain of thought prompting \cite{wei2023chainofthought}, which may be used to elicit truthful explanations of how models arrive at claims. However, \citet{turpin2023language} demonstrates that the given explanations can be misleading since models can be biased to change their answers in a way that is not reflected in their explanations--this is a form of explanation inconsistency.

There has been recent progress on this from work in interpretability. \citet{burns2022discovering} demonstrate that directions in the latent space of networks can be found that correspond to truthfulness better than the outputs of models directly. Our approach could complement techniques like this, providing new phenomena to better understand the trustworthiness of models.


A related investigation is into how language models respond to open-ended questions for which a single correct answer does not exist \cite{yin2023large}. Our work can be seen as considering the related case where instead of there being no correct answer, there exist multiple possible correct answers. Similarly, \citet{raj2023measuring, elazar-etal-2021-measuring} have focused on cross-prompt consistency over knowledge-focused QA.

\section{Conclusion}


All tested models behaved more self-consistently across contexts for ambiguous tasks than expected if the models had randomly consistent behaviour. This is surprising given models are not explicitly trained for cross-context self-consistency.
We also found that model consistency grows with model capability. We varied the task prompt, as well as the difficulty of the task (using base-2 sequences instead of base-10 sequences and varying the sequence length), and found that our findings are robust with respect to these changes. Across all evaluated models found that they are \textit{not} well calibrated when it comes to evaluating their own consistency.
We also tested that even when a model that chooses relatively consistent answers among several correct answers across contexts, models may still assign non-trivial probability to other correct answers. Asking the models to verbalize correct alternatives revealed high precision scores for all models which discern between correct and incorrect answers. In comparison, recall was relatively low where \texttt{text-davinci-003} surprisingly achieved the highest recall, closely followed by \texttt{gpt-4}, indicating they can retrieve alternative correct answers. The significance of our results is that we shouldn't assume the apparent consistency of LLMs points to actual internal consistency due to high probability mass placed on alternative answers which may equally be picked using common sampling techniques for natural language generation. As a community we should also be wary of consistency given our results on calibration that show models across capability classes strongly over and under estimate their own consistency.

\section{Limitations}

Ambiguous integer sequences is an idealized domain removing linguistic concerns and knowledge-related complexities of natural language tasks. Hence, results on this domain may not generalize. This is important because studies understanding LLMs safety typically focus on model behaviors that have a direct impact on understanding real-world risk, such as impact on socio-cultural prejudice or factual accuracy, of their deployment. Future work could investigate consistency in more general linguistic domains using a similar framework of ambiguity.

Our analysis of self-consistency was limited by only having access to models through a public API. In particular, we were only able to access the log probabilities of one model under analysis, and at the time of writing, this API is deprecated. Additionally, we did not include evaluation of available open-source models, which could have provided insightful comparisons with the OpenAI models and possibility to test output behaviours more extensively. Future work may be unable to access the log probabilities of these models to perform similar analyses. Although we did use greedy decoding with zero temperature, the GPT model tend to behave non-deterministically, which already introduces an implicit inconsistency and dependence on the few-shot examples. Reporting results averaged over several runs aimed to mitigate this. But controls for each experiments could have been done in addition to that. 
Our experiments in \ref{q2_alternative_considerations} were limited by the availability of token probabilities, so no scaling results are available in that section. We chose popular LLMs used through public APIs since we wanted to understand the behavior of those particular models, but future work should investigate open-source models that we are able to fully inspect. In particular, we believe the observed increase in cross-context consistency results from RLHF and pre-training. However, given the closed source nature of these models, it is possible that GPT-series models were trained with cross-context consistency objectives.

\section{Acknowledgements} Thanks to Julian Michael and Miles Turpin for feedback on a draft of this paper. This project has benefited from financial support to Sam Bowman by Eric and Wendy Schmidt (made by recommendation of the Schmidt Futures program) and Open Philanthropy. This material is based upon work supported by the National Science Foundation under Grant Nos. 1922658 and 2046556. Any opinions, findings, and conclusions or recommendations expressed in this material are those of the author(s) and do not necessarily reflect the views of the National Science Foundation. 



\bibliography{anthology,custom}
\bibliographystyle{acl_natbib}

\clearpage
\onecolumn
\appendix
\section{Mining Ambiguous Sequences}
\label{app:mining_ambigious_sequences}
\Cref{alg:mining_ambigious_sequences} describes how we find a set of ambiguous functions $\mathcal{A}$ given a set of function templates $\mathcal{F}_{templates}$ and the parameters $c:\mathbb{N}$ and $s:\mathbb{N}$ which control the sequence of constants to use for filling a set of templates and the number steps we must check that a pair of functions must match for.

\begin{definition}[Integer Sequence Ambiguity]
A pair of integer sequence-generating functions can be said to be ambiguous iff both functions generate the same sequence up to $|S|$ but generate different integers at step $|S| + 1$. This property holds if functions begin generation at different offsets.
\end{definition}

$\mathcal{F}_{templates}$ is a set of functions that have slots for constant terms used to construct the function space we will search for ambiguity within. For the purposes of our experiments, we generated templates using the function templates in \Cref{tab:function_templates} which consisted of templates with two constant term slots. We generated functions using integer constants in the range $[0, 4]$.

For our experiments, we checked ambiguity for sequences of length $4$ and an offset maximum of $4$. Unambiguous sequences are the complement of $\mathcal{A}$ and can easily be found by modifying the algorithm below to return sequences which are generated by only one function selected from the function space. It is important to note that the sequence is only unambiguous with respect to the function space selected.

\begin{table*}[h]
\centering
\begin{tabular}{ll}
\toprule
Type                          & Template                                                                            \\
\toprule
arithmetic progression        & \texttt{lambda x: (\{\} * x) + \{\}}                                                         \\ \midrule
geometric progression         & \texttt{lambda x: (\{\} * x) * \{\}}                                                       \\ \midrule
exponential progression       & \texttt{lambda x: (\{\} * x) ** \{\}}                                                       \\ \midrule
power progression             & \texttt{lambda x: \{\} ** (\{\} * x)}                                                        \\ \midrule
bit or progression            & \texttt{lambda x: (\{\} * x) | \{\}}                                                        \\ \midrule
modular progression           & \texttt{lambda x: (x * \{\}) \% (\{\}+1)}                                                    \\
\midrule
indexing criteria             & \texttt{lambda x:} \\
 progression                  & \texttt{ {[}i for i in range(100) if i \% (\{\} + 1) or i \% (\{\} + 1){]}{[}x{]}} \\
\midrule
recursive progression         & \texttt{(lambda a: lambda v: a(a,v))} \\
                              & \texttt{ (lambda fn,x: 1 if x==0 else \{\} * x * fn(fn,x-1) + \{\})} \\
\bottomrule
\end{tabular}
\caption{
\label{tab:function_templates}
Function templates with two constant term slots that were used for mining ambiguous sequences. Note our functions are indexed starting at one.
}
\end{table*}

\begin{algorithm*}
\caption{Mining Ambiguous Sequences}\label{alg:mining_ambigious_sequences}
\begin{algorithmic}
\Require{$\mathcal{F}_{templates}$} \Comment{Construct function space}
\State{Set $c \in C$ is a set of constants to parameterize the function templates}
\For{$f \in \mathcal{F}_{templates}$} 
    \For{$c_1 \in C$}
        \For{$c_2 \in C$}
            \State $\mathcal{F}_{filled} \gets \mathcal{F}_{filled} \cup {f[c1;c2]} $
        \EndFor
    \EndFor
\EndFor
\Require{$\mathcal{P} : \mathcal{F}_{filled} \times \mathcal{F}_{filled}$} \Comment{Check ambiguity}
\State{Set $S$ is a set of steps to check for ambiguity}
\State{Set $O$ us a set of offsets to check starting and ending positions}
\State{Set $\mathcal{A}$ is the set of ambiguous functions}
\For{$f_1, f_2 \in \mathcal{P}$} 
    \For{$o_1 \in O$}
        \For{$o_2 \in O$}
            \State{Set $seq_a$ is a temporary set for keeping track of the output from $f_1$}
            \State{Set $seq_b$ is a temporary set for keeping track of the output from $f_2$}
            \For{$s \in S$}
                \State{$seq_a \gets f_1(s + o_1)$}
                \State{$seq_b \gets f_2(s + o_2)$} 
                \If{$seq_a \neq seq_b$}
                    \State{break}
                \EndIf
                \If{$s = |S|$}
                    \State{$seq_a \gets f_1(s + o_1 + 1)$}
                    \State{$seq_b \gets f_2(s + o_2 + 1)$} 
                    \If{$seq_a \neq seq_b$}
                        \State{$\mathcal{A} \gets \mathcal{A} \cup {f_1, seq_a, f_2, seq_b}$}
                    \EndIf
                \EndIf
            \EndFor
        \EndFor
    \EndFor
\EndFor
\Ensure{Return $\mathcal{A}$ is the set of ambiguous functions identified above}
\end{algorithmic}
\end{algorithm*}

\clearpage
\section{Prompt Examples}
\label{app:prompt_examples}

\begin{figure*}[h]
\centering
\begin{lstlisting}[
    caption={System-level prompt. We prepend each context with minimal instruction and provide the valid space of lambda functions to the model. For base 2 sequences the outputs are wrapped with the bin function.},
    label={lst:prompt_system},
]
You are helping with integer sequences like arithmetic or geometric sequences. Sequences are not always 0 indexed, they may be offset by an arbitrary i-index value. Progressions are written as Python lambda functions and you may only respond in the same lambda-function format.
Valid lambdas:
- arithmetic -> lambda x: (a * x) + b
- geometric -> lambda x: (a * x) * b
- exponential -> lambda x: (a * x) ** b
- power -> lambda x: a ** (b * x)
- bitwise_or -> lambda x: (a * x) | b
- modular -> lambda x: (x * a) % (b+1)
- indexing_criteria -> lambda x: [i for i in range(100) if i % (a + 1) or i % (b + 1)][x]
- recursive -> (lambda a:lambda v:a(a,v))(lambda fn,x:1 if x==0 else a * x * fn(fn,x-1) + b)
\end{lstlisting}
\end{figure*}

\begin{figure*}[h]
\centering
\begin{lstlisting}[
    caption={Prompt for sequence explanation. We provide six few-shot examples before we prompt the model with the test prompt (highlighted in beige).},
    label={lst:prompt_sequence_explanation},
]
For the sequence: 3,7 
Give the code that generates the above sequence. 
lambda x: (2 * x) | 3 
For the sequence: 1,1,1 
Give the code that generates the above sequence. 
lambda x: (1 * x) ** 0 
For the sequence: 18,162 
Give the code that generates the above sequence. 
(lambda a:lambda v:a(a,v))(lambda fn,x:1 if x==0 else 3 * x * fn(fn,x-1) + 0) 
For the sequence: 4,7 
Give the code that generates the above sequence. 
lambda x: (3 * x) | 4
For the sequence: 1,1,1,1
Give the code that generates the above sequence. 
lambda x: 5 ** (0 * x) 
For the sequence: 3,5 
Give the code that generates the above sequence. 
lambda x: [i for i in range(100) if i % (3 + 1) or i % (3 + 1)][x] 
#\TESTPROMPT#For the sequence: 4,5 
#\TESTPROMPT#Give the code that generates the above sequence.
\end{lstlisting}
\end{figure*}

\begin{figure*}[h]
\centering
\begin{lstlisting}[
    caption={Prompt for sequence completion. We provide eight few-shot examples before we prompt the model with the test prompt (highlighted in beige).},
    label={lst:prompt_sequence_completion},
]
For the sequence: 2,3,4 
Complete the next number and only the next number. 
5 
For the sequence: 0,1,2,3 
Complete the next number and only the next number. 
0 
For the sequence: 0,0,0 
Complete the next number and only the next number. 
0 
For the sequence: 9,12,15,18
Complete the next number and only the next number. 
21 
For the sequence: 4,6,7,8 
Complete the next number and only the next number. 
9 
For the sequence: 3,8,26 
Complete the next number and only the next number. 
106 
For the sequence: 4,5,7 
Complete the next number and only the next number. 
8 
For the sequence: 2,6,24 
Complete the next number and only the next number. 
120 
#\TESTPROMPT#For the sequence: 12,15 
#\TESTPROMPT#Complete the next number and only the next number. 
\end{lstlisting}
\end{figure*}

\begin{figure*}[h]
\centering
\begin{lstlisting}[
    caption={Prompt for sequence explanation. We provide six few-shot examples before we prompt the model with the test prompt (highlighted in beige).},
    label={lst:prompt_sequence_explanation},
]
For the sequence: 3,7 
Give the code that generates the above sequence. 
lambda x: (2 * x) | 3 
For the sequence: 1,1,1 
Give the code that generates the above sequence. 
lambda x: (1 * x) ** 0 
For the sequence: 18,162 
Give the code that generates the above sequence. 
(lambda a:lambda v:a(a,v))(lambda fn,x:1 if x==0 else 3 * x * fn(fn,x-1) + 0) 
For the sequence: 4,7 
Give the code that generates the above sequence. 
lambda x: (3 * x) | 4
For the sequence: 1,1,1,1
Give the code that generates the above sequence. 
lambda x: 5 ** (0 * x) 
For the sequence: 3,5 
Give the code that generates the above sequence. 
lambda x: [i for i in range(100) if i % (3 + 1) or i % (3 + 1)][x] 
#\TESTPROMPT#For the sequence: 4,5 
#\TESTPROMPT#Give the code that generates the above sequence.
\end{lstlisting}
\end{figure*}

\begin{figure*}[h]
\centering
\begin{lstlisting}[
    caption={Prompt for self-consistency. We provide six few-shot examples before we prompt the model with the test prompt (highlighted in beige).},
    label={lst:prompt_self_consistency},
]
Is the following sequence: 1, 2, 3 consistent with the function lambda x: x + 1? 
Answer (Y/N): Y 
Is the following sequence: 3, 3, 4 consistent with the function lambda x: 3 * x + 2? 
Answer (Y/N): N 
Is the following sequence: 8, 18, 32 consistent with the function lambda x: 2 * x ** 2? 
Answer (Y/N): Y 
Is the following sequence: 256, 1024 consistent with the function lambda x: 4 ** x? 
Answer (Y/N): Y 
Is the following sequence: 7, 5, 12 consistent with the function lambda x: (3 * x) | 4? 
Answer (Y/N): N 
Is the following sequence: 2, 3, 0, 1 consistent with the function lambda x: (x * 5) % 4? 
Answer (Y/N): Y 
#\TESTPROMPT#Is the following sequence: 2,3,4 consistent with the function lambda x: (2 * x) | 0? 
#\TESTPROMPT#Answer (Y/N): 
\end{lstlisting}
\end{figure*}

\begin{figure*}[h]
\centering
\begin{lstlisting}[
    caption={Random prompt for explanations (preceded with few-shot examples). Note that "Sequences are not always 0 indexed , they may be offset by an arbitrary i-index value" from Listing 1 is omitted for Listings 6-11. },
    label={lst:prompt_random_explanation},
]
Pick a function which generates the following sequence.
If there are multiple possible functions, choose randomly.
Assume the first number was generated by f(0), the second by f(1), and so on.
The function should output numbers in the same representation as the sequence.
Give the response in the format:
Explanation: <function>
\end{lstlisting}
\end{figure*}
    
\begin{figure*}[h]
\centering
\begin{lstlisting}[
    caption={Random prompt for continuations (preceded with few-shot examples).},
    label={lst:prompt_random_continuation},
]
Choose a continuation of the following sequence. 
Assume the sequence is generated by some deterministic function. 
If multiple functions could generate the sequence, choose the corresponding continuation randomly.
\end{lstlisting}
\end{figure*}
    
\begin{figure*}[h]
\centering
\begin{lstlisting}[
    caption={Self-consistency prompt for explanations (preceded with few-shot examples).},
    label={lst:prompt_self_consistency_explanation},
]
Choose a function which generates the following sequence. 
Note that you will be separately asked to choose a continuation of the sequence. 
Please choose a function which is consistent with the continuation you will choose. 
Assume the first number was generated by f(0), the second by f(1), and so on. 
The function should output numbers in the same representation as the sequence. 
Give the response in the format: 
Explanation: <function>
\end{lstlisting}
\end{figure*}

\begin{figure*}[h]
\centering
\begin{lstlisting}[
    caption={Self-consistency prompt for continuations (preceded with few-shot examples).},
    label={lst:prompt_self_consistency_continuation},
]
Choose a continuation of the following sequence. 
Note that you will be separately asked to choose a function which generated the sequence. 
Please choose a continuation which is consistent with the function you will choose. 
Give the response as a number (with a base prefix if appropriate), with no additional commentary.
\end{lstlisting}
\end{figure*}
    
\begin{figure*}[h]
\centering
\begin{lstlisting}[
    caption={Most-likely prompt for explanations (preceded with few-shot examples).},
    label={lst:prompt_most_likely_explanation},
]
What is the most likely function which generated the following sequence? 
Assume the first number was generated by f(0), the second by f(1), the third by f(2), and so on. 
The function should output the number in the same representation as the sequence. 
Give the response in the format: 
Explanation: <function> 
\end{lstlisting}
\end{figure*}

\begin{figure*}[h]
\centering
\begin{lstlisting}[
    caption={Most-likely prompt for continuations (preceded with few-shot examples).},
    label={lst:prompt_most_likely_continuation},
]
What is the most likely continuation of the following sequence? 
Give the response as a number (with a base prefix if appropriate), with no additional commentary.
\end{lstlisting}
\end{figure*}
    
\begin{figure*}[h]
\centering
\begin{lstlisting}[
    caption={Prompt for verbalizing alternative completions.},
    label={lst:prompt_verbalize_alternative_continuations},
]
For the sequence: 3,7 
Complete the next possible number. 
Consider up to 5 possible and valid answers separated by escape character '\n', as determined by you, {model_name}. 
4 \n 15 \n 

#\TESTPROMPT#For the sequence: 1,1,1 
#\TESTPROMPT#Complete the next number and only the next number. 
#\TESTPROMPT#Consider up to 5 possible and valid answers separated by escape character '\n', as determined by you, {model_name}. 
\end{lstlisting}
\end{figure*}

\clearpage
\section{Robustness Experiment Results}
\label{sec:robustness-results}

\begin{table*}[h]
\centering
\small
\begin{tabular}{lll|ll|ll}
\hline
Model            & Base & Length & Correct & & Incorrect & \\ 
                 &      &        & Consistent & Inconsistent & Consistent & Inconsistent \\ \hline
gpt-4            & 10   & 4      & 70  & 2 & 2 & 26 \\
gpt-4            & 10   & 2      & 88  & 3 & 0 & 9 \\
gpt-4            & 2    & 4      & 23  & 4 & 2 & 72 \\
gpt-4            & 2    & 2      & 19  & 15 & 0 & 66 \\
gpt-3.5-turbo    & 10   & 4      & 65  & 4 & 2 & 26 \\
gpt-3.5-turbo    & 10   & 2      & 38  & 16 & 2 & 44 \\
gpt-3.5-turbo    & 2    & 4      & 11 & 2 & 0 & 84 \\
gpt-3.5-turbo    & 2    & 2      & 9 & 3 & 2 & 81 \\

\end{tabular}
\caption{The proportion of self-consistent continuation and explanation pairs (Consistent), alongside whether the explanations are correct (Correct), for a given model (Model) on generated ambiguous sequences of length (Length), represented in base (Base). Also tracks whether explanations or continuations are invalid (Invalid).}
\end{table*}

\section{Histogram of Log Probabilities for Alternative Completions of Ambiguous Sequences.}
\label{sec:app-alternatives}

\begin{figure}[h]
    \centering
    \includegraphics[scale=0.5]{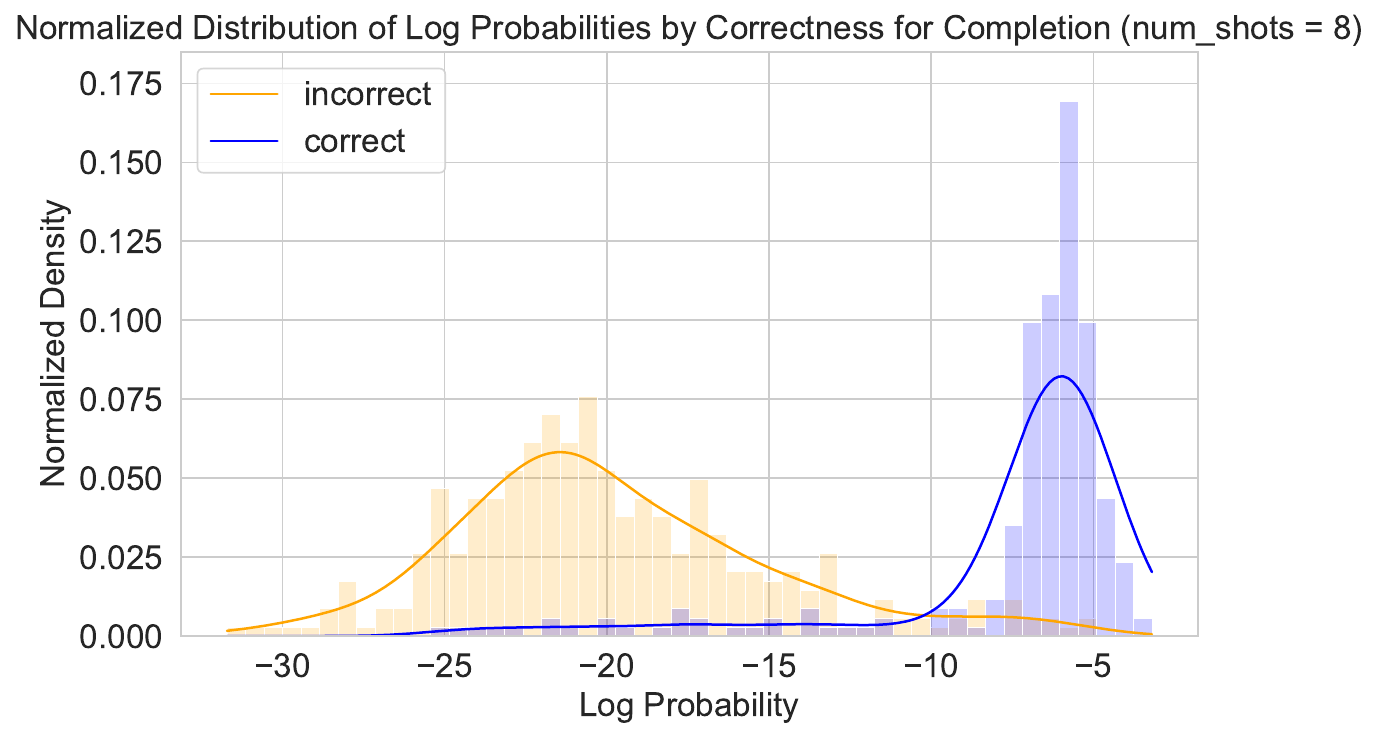}
    \caption{Distribution over log probabilities by correctness with densities normalized across all data points. It shows a narrow concentration of relatively large probabilities for correct answers and incorrect answers with relatively small probabilities. The plot shows results for few-shots examples of \texttt{random} samples; distributions with different few-shot sampling methods and number of shots look very similar.}
    \label{fig:app-prob-mass-normalized}
\end{figure}


\end{document}